\documentclass[journal]{IEEEtran}

\usepackage{ifpdf}
\usepackage{cite}

\ifCLASSINFOpdf
  \usepackage[pdftex]{graphicx}
  \graphicspath{{./figures/}}
  \DeclareGraphicsExtensions{.pdf,.jpeg,.png}
\else
  \usepackage[dvips]{graphicx}
  \graphicspath{{./figures/}}
  \DeclareGraphicsExtensions{.eps}
\fi

\usepackage{amsmath}
\usepackage{algorithmic}
\usepackage{array}

\ifCLASSOPTIONcompsoc
  \usepackage[caption=false,font=normalsize,labelfont=sf,textfont=sf]{subfig}
\else
  \usepackage[caption=false,font=footnotesize]{subfig}
\fi

\usepackage{stfloats}
\usepackage{url}
\usepackage{booktabs}

\begin{document}

\title{StarCraft II Build Order Optimization using Deep Reinforcement Learning and Monte-Carlo Tree Search}

\author{Islam~Elnabarawy,~\IEEEmembership{Student~Member,~IEEE,}
        Kristijana~Arroyo,~\IEEEmembership{Stduent~Member,~IEEE,}
        Donald~C.~Wunsch~II,~\IEEEmembership{Fellow,~IEEE}%
    \thanks{Islam Elnabarawy, Kristijana Arroyo, and Donald C. Wunsch II are with the Applied Computational Intelligence Laboratory at the Missouri University of Science and Technology.}%
    \thanks{
        Islam Elnabarawy and Kristijana Arroyo are with the Department of Computer Science, Missouri University of Science and Technology, Rolla, Missouri 65409. E-mail: elnabarawy@ieee.org, ka436@mst.edu.}%
    \thanks{
        D. Wunsch is with the Department of Electrical and Computer Engineering, Missouri University of Science and Technology, Rolla, Missouri 65409. E-mail: wunsch@ieee.org.}%
}

\maketitle

\begin{abstract}
The real-time strategy game of StarCraft II has been posed as a challenge for reinforcement learning by Google's DeepMind. This study examines the use of an agent based on the Monte-Carlo Tree Search algorithm for optimizing the build order in StarCraft II, and discusses how its performance can be improved even further by combining it with a deep reinforcement learning neural network. The experimental results accomplished using Monte-Carlo Tree Search achieves a score similar to a novice human player by only using very limited time and computational resources, which paves the way to achieving scores comparable to those of a human expert by combining it with the use of deep reinforcement learning.
\end{abstract}


\IEEEpeerreviewmaketitle

\section{Introduction}

\IEEEPARstart{V}{ideo} games often provide a challenging and innovative platform for developing and improving artificial intelligence and machine learning algorithms, particularly in the area of reinforcement learning. The game of StarCraft is a popular example of this, and has been fueling the development of intelligent agents for over two decades \cite{6637024}. StarCraft is a real-time strategy video game where players are responsible for gathering resources, building technology, recruiting an army, and defeating their opponents in combat. Both StarCraft\textsuperscript{\textregistered}\footnote{\textcopyright1998 Blizzard Entertainment, Inc. All rights reserved.} and StarCraft\textsuperscript{\textregistered} II\footnote{\textcopyright2010 Blizzard Entertainment, Inc. All rights reserved.}\footnote{StarCraft and Blizzard Entertainment are trademarks or registered trademarks of Blizzard Entertainment, Inc. in the U.S. and/or other countries.} still enjoy widespread popularity despite their age, and there are professional competitions and world championships actively being held for StarCraft II, with the winners receiving large monetary rewards and major sponsorship deals. World champion players often demonstrate a high level of strategic control as well as extremely rapid reflexes and multitasking abilities.

Google's DeepMind issued a challenge using StarCraft II as a platform for developing intelligent agents that can defeat world champion human players \cite{Vinyals2017,Vinyals2019}. To assist the research and development of these agents, they collaborated with Blizzard Entertainment to release an Application Programming Interface (API) for StarCraft II, which facilitates the development of agents that can play the game programmatically. This was accompanied by the release of the StarCraft II Learning Enviroment (SC2LE) by DeepMind, which contains an open source reinforcement learning environment called PySC2, as well as a set of anonymized replays from human matches and a set of mini-games intended as small test scenarios for reinforcement learning agents.

\begin{figure}
	\centering
	\includegraphics[width=0.9\linewidth]{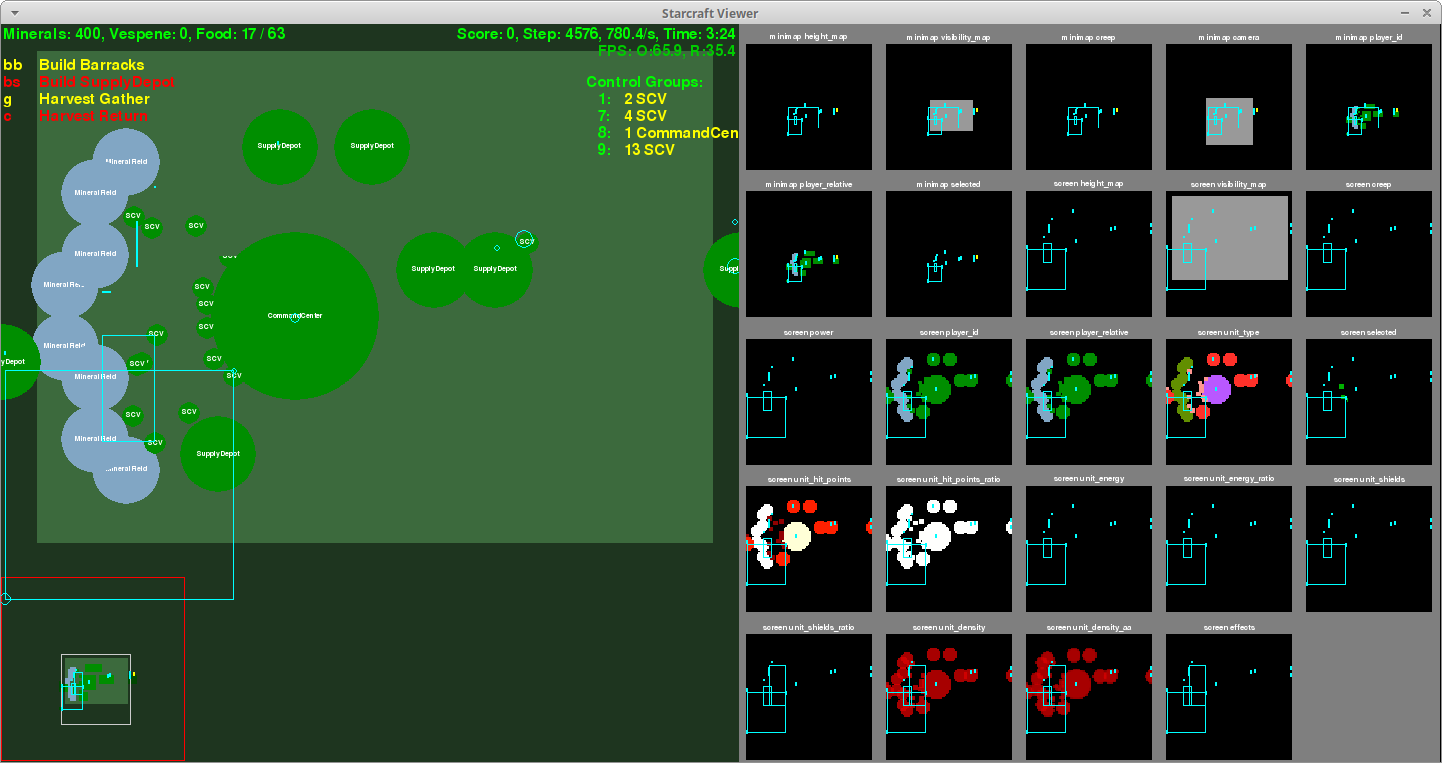}
	\caption{An example of the spatial interface in SC2LE as seen through the PySC2 environment}
	\label{fig:pysc2gui}
\end{figure}

\begin{figure}
	\centering
	\includegraphics[width=0.9\linewidth]{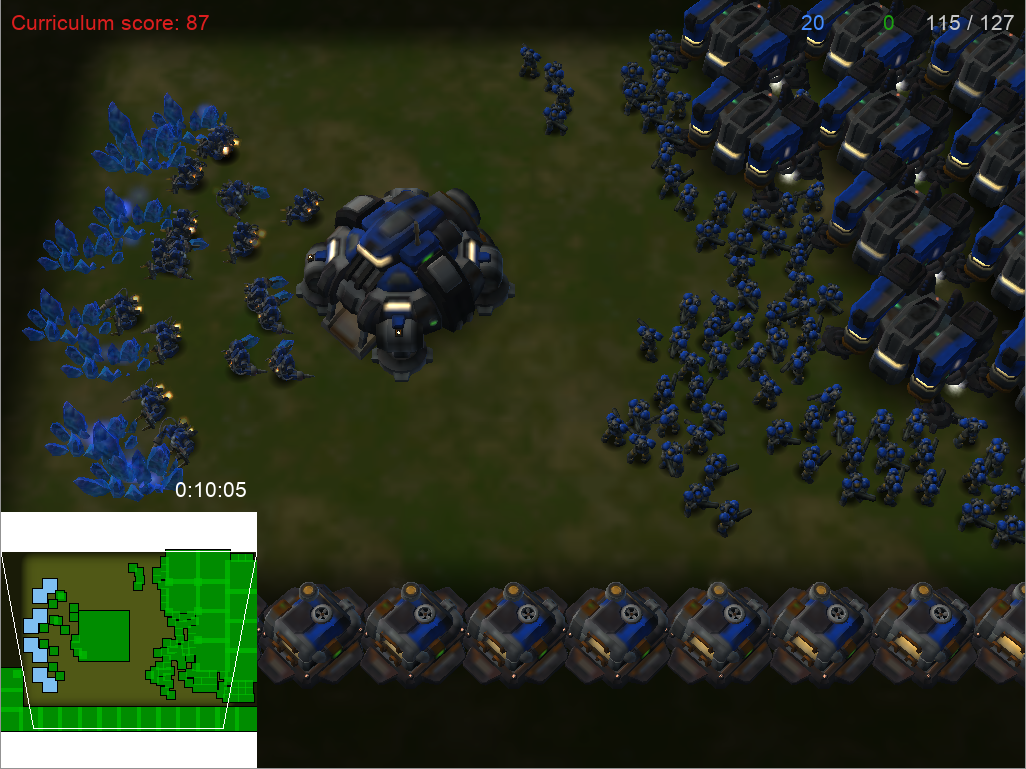}
	\caption{An example of the render interface of SC2LE}
	\label{fig:buildmarinesvisualizer}
\end{figure}

The StarCraft II reinforcement learning environment consists of observations, actions, and rewards. The observations can be in one or more of three possible formats: raw, spatial, or RGB. The raw observations consist of the complete data associated with each of the units on the map, including neutral units such as resources, units controlled by the agent, and each of the opponent's units that the agent can currently observe. The spatial observations consist of a number of 2-dimensional layers of data that each relay some aspect of the map that the agent can currently observe, such as which points are occupied by units, what type each unit is, and who the owner of each unit is. Each layer is an orthogonal projection of the map conveying some type of information, presented in 2-dimensional space corresponding with the map. The RGB observation is a true render of what the game looks like to a human player, and is therefore often referred to as the render interface. An RGB observation consists of a single 2-D image of the main game screen, and a single 2-D image representing the minimap (birds eye) view. Fig.~\ref{fig:pysc2gui} shows an example of the spatial interface observations, while Fig.~\ref{fig:buildmarinesvisualizer} shows an example of the render interface observations.

In a typical game of StarCraft II, each player has to focus on two levels of gameplay: individual unit movement, often referred to as ``micro'', and overall strategy, often referred to as ``macro''. For a player to be able to achieve a strong position, they typically have to control and monitor their resource gathering units, use the resources they've gathered to create buildings and upgrade their technology, as well as amassing their army to defend their base and eventually attack their opponent's base. However, since resources are limited and can only be collected at a certain rate, the player has to strategically plan which technology to invest in, which types of units to build, and how many of each type. Human players often follow certain patterns in their macro strategies, and continue to test and refine them. These patterns of macro strategy are known as a build order, and can be though of as similar to an opening in the game of Chess.

In this study, we focus on exploring the challenge of macro strategy in StarCraft II through the discovery and optimization of build order patterns, with the aim of building an agent that can continuously optimize and adapt its build order during gameplay based on its observation of the state of the game and the opponent's actions. We accomplish this through the combination of the Monte-Carlo Tree Search algorithm in combination with a deep reinforcement learning neural network, an approach whose effectiveness was demonstrated through AlphaGo \cite{Silver2017a,Silver2016}, which was able to defeat the world champion in the game of Go, and AlphaZero \cite{Silver2017}, which showed superior gameplay in Chess and Shogi.

To limit the scope of this initial proof of concept, this study uses one of the minigames developed by DeepMind and which limits the agent to solving the problem of build order. The minigame is called ``BuildMarines'', and the objective is for the agent to gather resources and build technology that allows it to produce as many infantry units, or ``Marines'', as possible during a fixed amount of time. The advantage of using this minigame is that it limits the number of possible actions and the possible technology to consider, while still requiring the agent to master the balance between gathering resources and producing buildings and units. This allows the agent to be trained within reasonable limits in time and computational power, while allowing the scope to be extended to the full game by allowing the agent access to more computational time and resources.
 
The remainder of this paper is organized as follows. Section~\ref{sec:background} presents some background concepts required for understanding the main contribution of this study. Section~\ref{sec:methodology} explains the approach used in the experiments, and Section~\ref{sec:experiment} provides the details of the experimental setup. Section~\ref{sec:results} discusses the results of the experiments, followed by this paper's conclusion in Section~\ref{sec:conclusion}.
 
\section{Monte-Carlo Tree Search}\label{sec:background}

Monte-Carlo Tree Search (MCTS) \cite{Browne2012} is an optimization method that finds the optimal decisions for a given problem by constructing a search tree via a random sampling of the decision space. It has been widely used in many application areas with great success, including in 2-player turn-based perfect information games.

The fundamental MCTS algorithm follows a simple 4-step process:
\begin{enumerate}
	\item Selection
	\item Expansion
	\item Simulation
	\item Backpropagation
\end{enumerate}

The algorithm starts from the current state and builds a tree representing all the possible decisions or actions from each state. The selection stage begins by moving down the tree of already discovered states, as long as all the possible actions for each state have been explored previously. The choice of which node to select is governed by a selection strategy, which will be discussed later. This continues until the search reaches a state in which not all the actions have yet been explored in the current tree. Then, the expansion stage randomly chooses one of the unexplored actions and expands it to discover and record the new state.

After that, the simulation stage conducts a Monte-Carlo simulation of randomly chosen actions from the current state, until it reaches a terminal state. For example, in a game, it would randomly choose from the legal moves and continue doing so until the game concludes in either a win or a loss. After that, the backpropagation stage goes back up the tree, starting from the node where the simulation began, and updates each node's simulation count, as well as incrementing the node's win count if the simulation ended in a win for the player corresponding to that node's decision. This backpropagation continues up the tree until the node which started the search process.

There are several strategies for deciding which node to select during the selection stage. The Upper Confidence Bound (UCB1) method looks at the confidence intervals for each action, and balances the exploitation of known good states with the exploration of possibly more optimal states. The value of each of the current node's children is calculated as:
\begin{equation}
\dfrac{x_i}{n_i} + \beta \sqrt{\frac{2 \ln n}{n_i}}
\end{equation}
where $x_i$ is the total number of wins for the child node, $n_i$ is the total number of visits for the child node, $n$ is the total number of visits for the current state, and $\beta \in (0, 1] $ is a parameter controlling the level of exploration vs. exploitation. The UCB1 strategy selects the node with the highest value at each step.

\section{Methodology}\label{sec:methodology}

In this study, the raw interface was used to build an abstract state representation based on the ``BuildMarines'' minigame. At each game step, with the game stepping through approximately 3 times per second, the current state of the game is collected from the raw interface and then converted into the following abstract state representation:
\begin{itemize}
    \item $w$: total number of workers
    \item $r$: number of barracks
    \item $b$: number of bases
    \item $p$: number of marines
    \item $s$: amount of supply remaining
    \item $m$: amount of minerals available
    \item $c$: supply capacity
    \item $t$: time
    \item $w_t$: time until a worker is produced
    \item $s_t$: time until a supply depot is produced
    \item $b_t$: time until a barracks is produced
    \item $p_t$: time until a marine is produced
\end{itemize}

The MCTS search algorithm is invoked using this abstract state representation, and the next state is simulated by updating the appropriate state variable based on the selected action. The set of available actions at each state are:
\begin{itemize}
	\item \textit{no-op}
	\item \textit{train worker}
	\item \textit{build supply}
	\item \textit{build barracks}
	\item \textit{train marine}
\end{itemize}

In addition to updating the variables based on the chosen action at each state, production queue times are advanced each frame and mineral collection is simulated with the passage of time at an estimated mineral collection rate based on the number of workers. The game is concluded at the end of the allocated time interval of approximately 10 minutes, and the final episode score is taken as the number of marines present in the final state.

\section{Experimental Setup}\label{sec:experiment}

To conduct an evaluation of the performance of the MCTS algorithm on build order optimization in StarCraft II, an agent was developed using the SC2LE environment. The agent was set to use the UCB1 selection strategy with an exploration parameter value of $\beta = \dfrac{1}{\sqrt{2}}$, which was chosen as a reasonable default based on recommendations from the MCTS literature.

The experiment was run multiple times, allowing the MCTS algorithm to do progressively more iterations of work at each time step. Since the BuildMarines minigame has little room for randomness and StarCraft II is a deterministic environment contingent on the pseudo-random number generator seed, the experiment was only conducted once for each iteration count. The final score were collected at the end of each experimental run, indicated by the total number of marines present at the end of the simulation.

\section{Results and Discussion}\label{sec:results}

\begin{figure}
	\centering
	\includegraphics[width=0.9\linewidth]{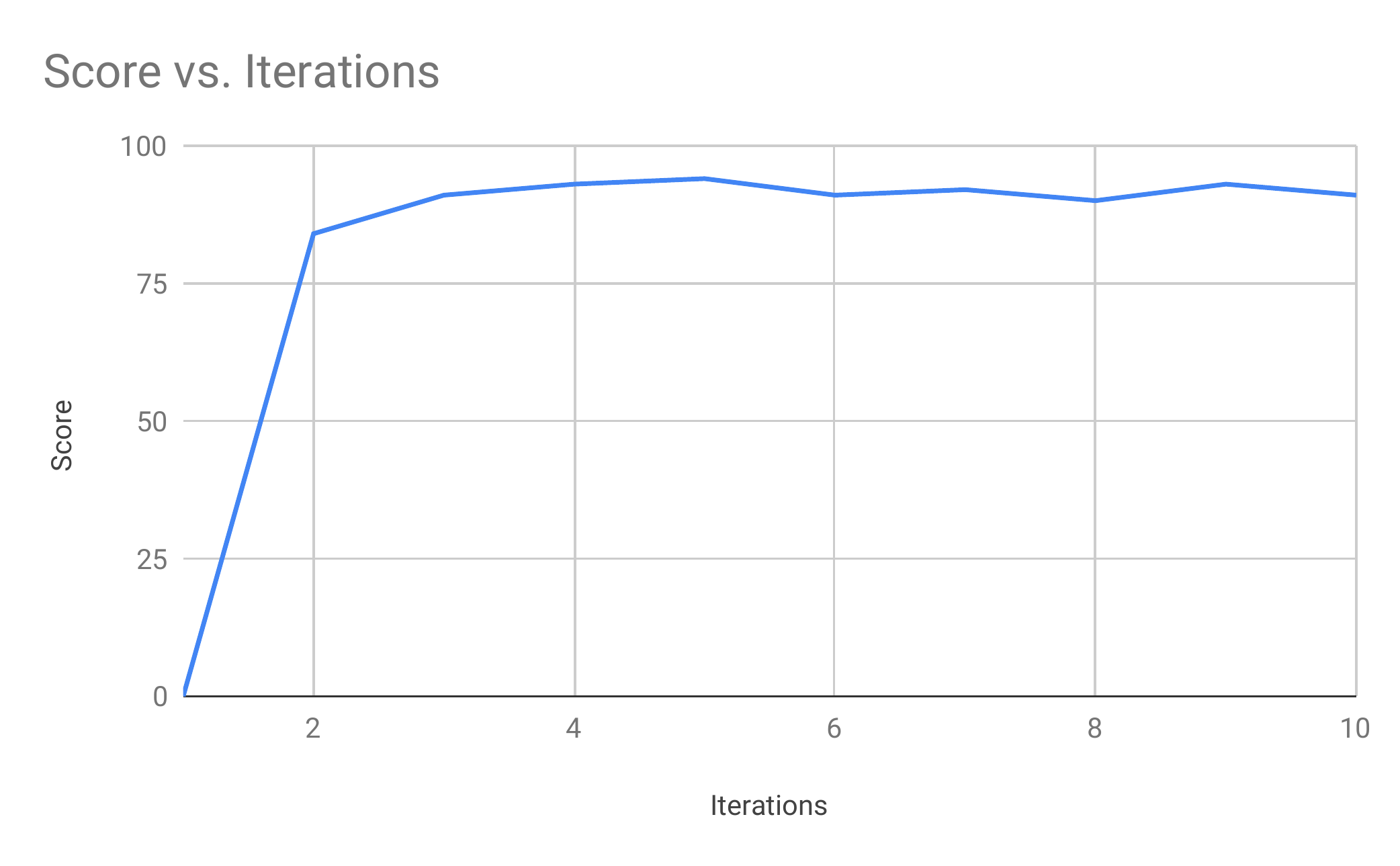}
	\caption{Line plot showing the MCTS agent score across the different iteration limits}
	\label{fig:results}
\end{figure}

Table~\ref{tab:results} shows the results of running the MCTS agent with iteration limits between 1 and 10. The same results are illustrated in Fig.~\ref{fig:results}.

\begin{table}[htpb]
	\centering
	\caption{Results of running the MCTS agent on the BuildMarines minigame with different iteration limits}
	\label{tab:results}
	\begin{tabular}{@{}cc@{}}
		\toprule
		\textbf{Iterations} & \textbf{Score} \\ \midrule
		1                   & 0              \\
		2                   & 84             \\
		3                   & 91             \\
		4                   & 93             \\
		5                   & 94             \\
		6                   & 91             \\
		7                   & 92             \\
		8                   & 90             \\
		9                   & 93             \\
		10                  & 91             \\ \bottomrule
	\end{tabular}
\end{table}

The scores achieved by the agent were comparable to those achieved by a novice human player (see \cite{Vinyals2017}) but below those achievable by an experienced human player. Moreover, the score is not affected by increasing the MCTS iteration limit within this range beyond the expected fluctuations due to the randomness in MCTS rollouts. It is possible to achieve scores over 100, but only by allowing the algorithm significantly more time to run by increasing the iteration limit by an order of magnitude.

The results illustrate the viability of this approach for optimizing the build order, where the simple MCTS agent can achieve reasonable results directly and with very little computational resources. The addition of a deep reinforcement learning neural network is therefore anticipated to improve these results significantly by reducing the amount of time needed for evaluating each iteration. Instead of doing costly rollouts, the neural network would be used to estimate the value of the current state, allowing the MCTS algorithm to do a much more exhaustive search within the same computational constraints.

\section{Conclusion}\label{sec:conclusion}

This study presents an initial exploration into the use of the Monte-Carlo Tree Search algorithm in combination with a deep reinforcement learning neural network for the optimization of build order in the game of StarCraft II. It presents an initial experiment limited to the use of the MCTS algorithm in designing a build order optimization agent, and immediately shows performance comparable to a novice human player. The limitations of computing time restrict the MCTS agent from performing closer to human experts, which is anticipated to be achievable through the combination of MCTS and a deep reinforcement learning neural network.

\section*{Acknowledgment}

Support from the Missouri University of Science and Technology Intelligent Systems Center and the Mary K. Finley Missouri Endowment is gratefully acknowledged.

This research was sponsored by the Army Research Laboratory (ARL) and the Lifelong Learning Machines program from DARPA/MTO, and it was accomplished under Cooperative Agreement Number W911NF-18-2-0260. The views and conclusions contained in this document are those of the authors and should not be interpreted as representing the official policies, either expressed or implied, of the Army Research Laboratory or the U.S. Government. The U.S. Government is authorized to reproduce and distribute reprints for Government purposes notwithstanding any copyright notation herein.


\bibliographystyle{IEEEtran}
\bibliography{bibliography}

\end{document}